\documentclass[letterpaper, 10 pt, conference]{ieeeconf}  

\IEEEoverridecommandlockouts                              

\usepackage{amsmath} 
\usepackage{amssymb}  
\usepackage{array,multirow,graphicx}
\usepackage{makecell}
\usepackage{subcaption}
\usepackage{algorithmicx}
\usepackage{algorithm}
\usepackage{algpseudocode}
\usepackage{booktabs}
\usepackage[bookmarks=true]{hyperref}
\usepackage{wrapfig,lipsum,booktabs}

\usepackage{siunitx}

\newcount\Comments 
\usepackage{array,multirow,graphicx}
\usepackage{color}
\usepackage{wrapfig}
\definecolor{darkgreen}{rgb}{0,0.5,0}
\definecolor{purple}{rgb}{1,0,1}
\newcommand{\kibitz}[2]{\ifnum\Comments=1\textcolor{#1}{#2}\fi}

\title{\LARGE \bf
Dynamic Grasping with Reachability and Motion Awareness
}

\author{
Iretiayo Akinola\thanks{$^*$Equal Contribution}$^{*}$, Jingxi Xu$^{*}$, Shuran Song and Peter K. Allen \\ 
\thanks{ 
        This work was supported in part by National Science Foundation grants IIS-1734557 and IIS-1527747. Authors are with the Department of Computer Science, 
        Columbia University, New York, NY 10027, USA.
        {\tt\small  (iakinola, jxu, shurans, allen)@cs.columbia.edu},
}}

\begin{document}


\maketitle
\thispagestyle{empty}
\pagestyle{empty}


\begin{abstract}

Grasping in dynamic environments presents a unique set of challenges. A stable and reachable grasp can become unreachable and unstable as the target object moves, motion planning needs to be adaptive and in real time, the delay in computation makes prediction necessary. In this paper, we present a dynamic grasping framework that is reachability-aware and motion-aware. Specifically, we model the reachability space of the robot using a signed distance field which enables us to quickly screen unreachable grasps. Also, we train a neural network to predict the grasp quality conditioned on the current motion of the target. Using these as ranking functions, we quickly filter a large grasp database to a few grasps in real time. In addition, we present a seeding approach for arm motion generation that utilizes solution from previous time step. This quickly generates a new arm trajectory that is close to the previous plan and prevents fluctuation. We implement a recurrent neural network (RNN) for modelling and predicting the object motion. Our extensive experiments demonstrate the importance of each of these components and we validate our pipeline on a real robot.
\end{abstract}

\section{INTRODUCTION}
\begin{figure}[t]
\vspace{2mm}
\begin{center}
    \includegraphics[width=\linewidth]{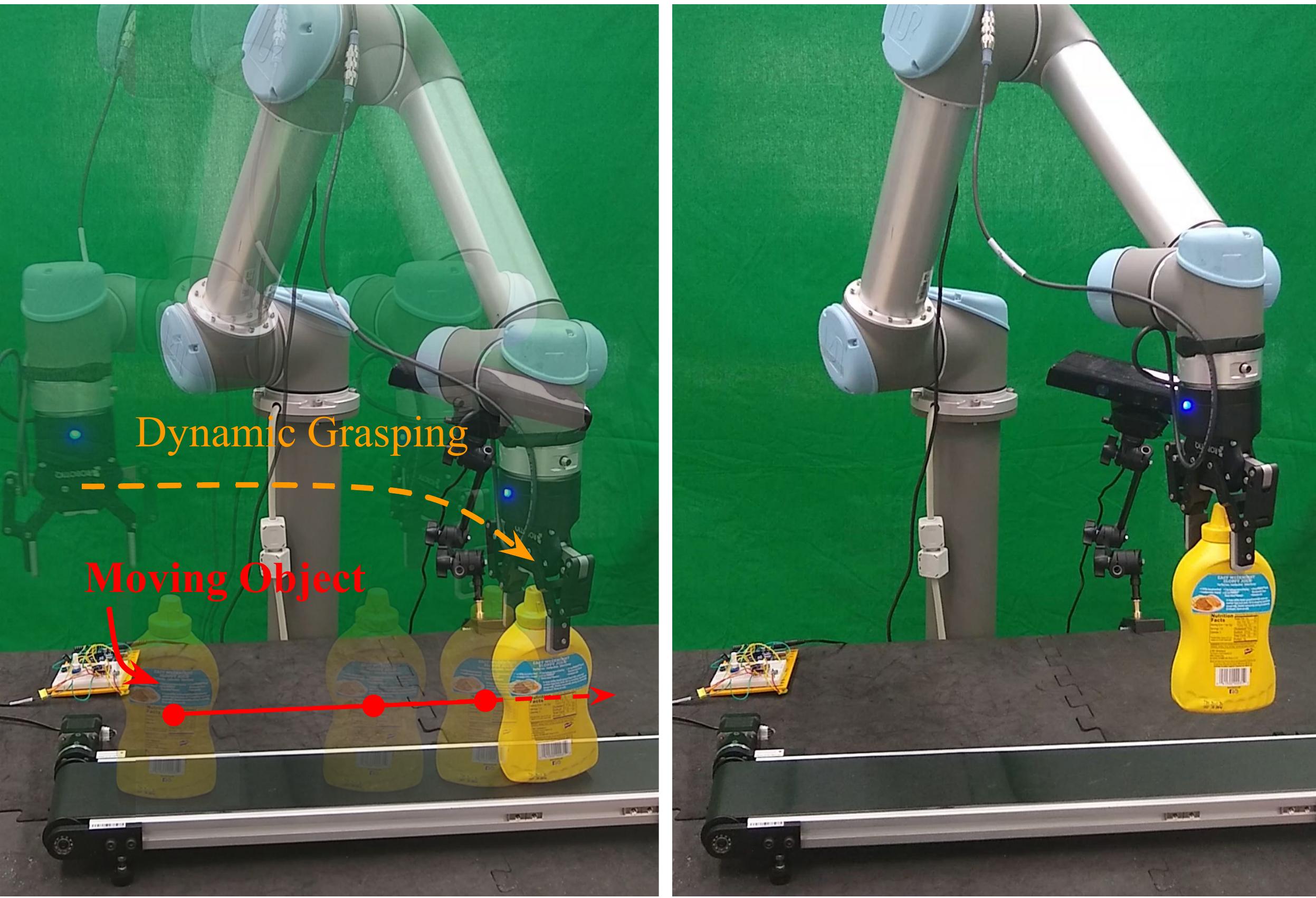}
\end{center}
\caption{\small Dynamic Grasping Problem: A moving target object is to be grasped and lifted. The object pose and motion is not known a priori and has to be estimated online. Full degree-of-freedom grasps should be explored to come up with feasible grasps that can pick-up the object before it escapes the robot's workspace.}
\label{fig:illustration}
\vspace{-6mm}
\end{figure}

Roboticists have made significant progress in developing algorithms and methods for robotic manipulation in static environments. However, robotic manipulation becomes much harder in dynamic environments which is often the case in the real world. For example, in dynamic grasping, ball catching, human-robot handover, etc., the targets and obstacles to be interacted with might be moving with an unknown motion. Providing robots with the ability to manipulate objects in dynamic environments, despite being less explored, can be extremely important in realizing automation in both industry and daily life. Figure~\ref{fig:illustration} illustrates a conveyor belt setting; an ability to pick up the target object without pausing the conveyor belt or knowing the speed of the target object a priori can improve the overall efficiency of the system.

There are many challenges brought by dynamic environments. 
First, continuous changes in the environments require online and fast motion replanning. Sampling-based methods (RRT, PRM, etc.) are not well-suited for this requirement because the randomness of solutions leads to jerky and wavy motion due to the replanning at each time step. Optimization-based methods (CHOMP, STOMP, etc.) can be time-consuming in highly cluttered scenes, making fast replanning in dynamic environments extremely difficult.
Second, most works in the grasp planning literature rarely consider the approach and close motion of the grasp, which makes a difference for a moving target. For example, a grasp facing the moving direction of a target can have a higher success rate than a grasp catching the target from the back. Third, we need to understand and predict the motion of the object because computed plans are obsolete when executed.


\begin{figure*}[t]
\vspace{2mm}
\begin{center}
    \includegraphics[width=0.245\textwidth]{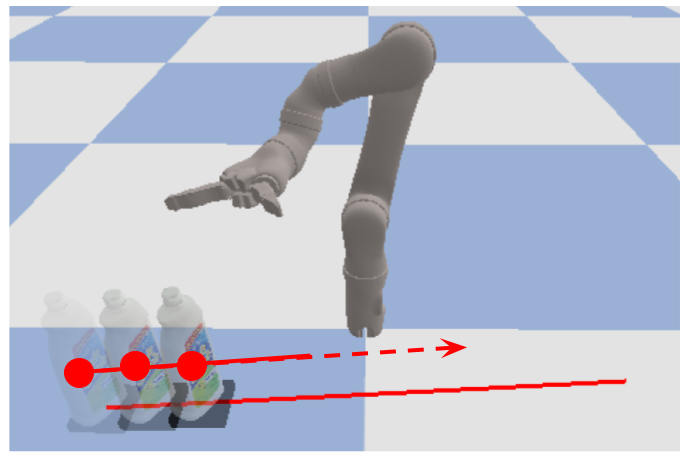}
    \includegraphics[width=0.245\textwidth]{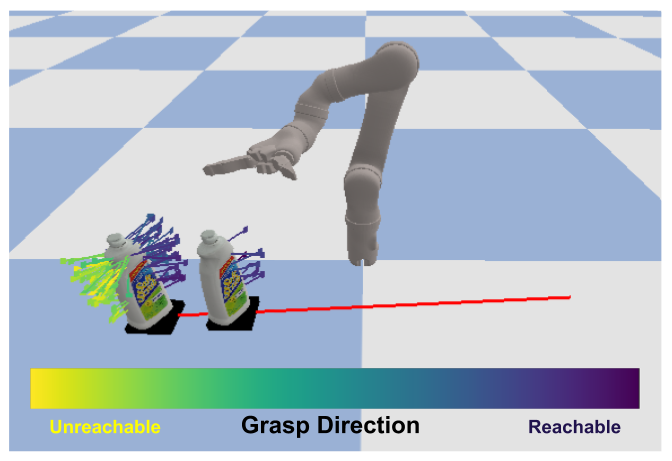}
    \includegraphics[width=0.245\textwidth]{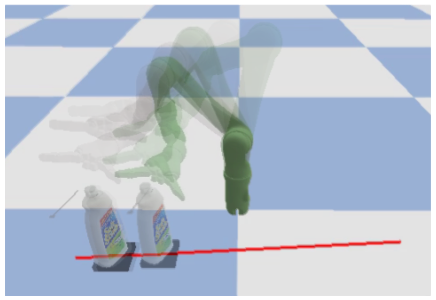}
    \includegraphics[width=0.245\textwidth]{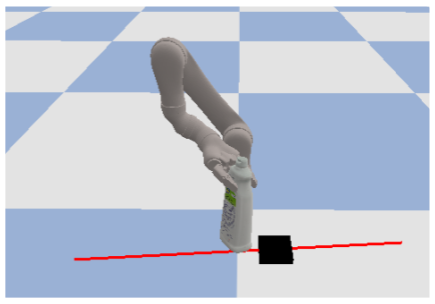}
    (a) Motion Prediction ~~~~~~~~~ (b) Grasp Planning ~~~~~~~~~~  (c) Trajectory Generation ~~~~~~~~~~~ (d) Grasp Execution
\vspace{-2mm}
\end{center}
\caption{ \small Dynamic Grasping Framework.
\textbf{a)} Instantaneous pose estimation runs continuously to keep track of the moving object and we use a recurrent neural network to model the motion of the target object and predict its future pose.
\textbf{b)} Full grasp database are ranked and filtered based on reachability.
\textbf{c)} Pick the grasp from filtered list that is closest to the current arm configuration. Arm trajectory is generated based on the future pose of the moving object. Arm trajectory from previous time step is used to seed the planner in current step. \textbf{d)} Approach and grasp are executed when \textsc{CanGrasp} condition is satisfied.}
\label{fig:full_pipeline}
\vspace{-4mm}
\end{figure*}

Previous works have addressed dynamic grasping by introducing a number of assumptions such as prior knowledge of the object motion \cite{allen1993automated}, waiting for the object to come to rest before grasping, limiting the grasping directions to a single direction (e.g. only top-down grasps \cite{ye2018velocity}).
In this work, we relax some of these assumptions and tackle the problem of robotic grasping for moving objects with no prior knowledge of the object's motion profile and no restrictions on the possible grasping directions of the object.
The increase in the range of possible grasping directions has the advantage of expanding the workspace of the robot leading to more grasp options that can be very useful in the dynamic setting. However, as the range of feasible grasp options grows, so does the range of infeasible ones. Without a notion of reachability, it is usually preemptively time-consuming to compute IKs for all the grasps in the database. Our method, illustrated in Figure \ref{fig:full_pipeline}, embraces the advantage of an expanded workspace for full degree-of-freedom (DOF) grasps and mitigates the reachability problem by constraining the grasp selection process to the more reachable and manipulable regions of the workspace. In addition, we observe that the robustness of a grasp may vary depending on the speed and direction of the moving object. To handle this, we learn a function that predicts the robustness of grasps given the motion of the object. This is used to rank and select a robust reachable grasp.
To generate arm motion, rather than planning from scratch each time, we seed the planning process by the solution from the previous time step. This method allows the newly planned trajectory to be similar and also speed up the computation. In summary, our main contributions are:

\paragraph{\textbf{Reachability and motion-aware grasp planning}} ranking functions that predict the reachability and success probability for different grasps based on target pose and motion of the target. These ranking functions are used for real-time grasp filtering. 
\paragraph{\textbf{Adaptive motion generation}} an effective trajectory generation approach that incorporates the solution from previous timestep to seed the search process to achieve quicker and smoother transition between different motion plans. 
\paragraph{\textbf{Simulation and real robot evaluation}} procedure of systematically evaluating dynamic grasping performance in simulation with randomized linear\,/\,nonlinear motion with different objects, and a real robot demonstration to pick up objects moving on a conveyor belt. 


\section{RELATED WORKS}

\subsubsection{Grasping in Dynamic Environments}
\label{sec:dynamic_grasping_related_works}
Grasping of static objects can be achieved via visual servoing~\cite{kragic2002survey, vahrenkamp2008visual, kragic2003framework, houshangi1990control, wilson1996relative, vahrenkamp2009visual, mahony2008dynamic, xie2014research}; however, grasping in a changing environment presents a unique challenge.
The robot not only tracks the object but also has to reason about the geometry of the object to determine how to pick it up.
Some learning-based grasping systems \cite{morrison2018closing} have been applied to slightly moving scenes.
Previous works demonstrate grasping of a static object in the midst of moving obstacles~\cite{kappler2018real}.
Our work deals with picking a moving object while avoiding collision with static obstacles.

\subsubsection{Database-based Robotic Grasping}
Previous works \cite{goldfeder2009columbia, goldfeder2011data} have looked at the idea of grasping using a precomputed database. Most of these methods sample different grasps and evaluate them in simulation using a geometry-based metric. These metrics use static analysis which does not account for the dynamics of the approach and lift process.
Some other methods~\cite{lenz2015deep, pinto2016supersizing} generate grasp database using a real robot which can be valuable but such data is very expensive to collect. 
A recent concurrent work~\cite{eppner2019billion} used this technique to examine different approaches for sampling grasps when generating a grasp database and evaluated their coverage of possible grasping directions. 
They very densely sample ``billions'' of grasping direction and measure the robustness of each grasp candidate using the success rate of it's neighbours. ~\cite{weisz2012pose} collects grasps with randomly added perturbations on the object poses. 

\subsubsection{Object Tracking}
\label{sec:object_tracking_related_works}
Visual feedback is crucial to grasping and manipulation in dynamic environments. For a position-based system like ours, the visual input from a camera (color and/or depth) is continuously processed into the pose (position and orientation) of the objects in the environment. 
Bayesian methods~\cite{issac2016depth, wuthrich2013probabilistic} or deep learning techniques \cite{tremblay2018deep, wang2019densefusion} can be applied to the input image stream to produce object poses in the camera's frame of reference. The noisy pose results from object pose detection systems can be filtered into more stable values using methods such as Kalman filtering~\cite{kalman1960new}. Since Kalman filtering builds a model for the motion, this model serves as a good predictor for the future pose of the object being tracked. 

\subsubsection{Motion Generation}
\label{sec:arm_traj_related_works}

When obstacles are present, reaching a moving target requires some trajectory planning (such as RRT~\cite{kuffner2000rrt}, PRM~\cite{kavraki1996probabilistic}) or trajectory optimization methods (such as CHOMP~\cite{ratliff2009chomp}, STOMP~\cite{kalakrishnan2011stomp}) that are able to generate collision-free paths for the arm. 
Recent works \cite{schmitt2019planning, schmitt2019modeling} presented an approach to generate a sequence of constraint-based controllers to reactively execute a plan while respecting specified constraints like collision avoidance.
Our work is more similar to works that generate arm motion from a library of stored arm motions \cite{berenson2012robot, coleman2015experience, islam2021provably}. 
Building on these works, we propose an approach that only keeps the solution from previous time step without a precomputed database of arm motions. This previous solution is used to initialize tree/roadmap for sampling-based methods or as a seed for trajectory optimization solver. 



A recent work \cite{ye2018velocity} presented an approach that uses motion prediction to grasp a moving block using top-down grasps; they illustrated their approach using simulated experiments. Our work differs from \cite{ye2018velocity} in that we do not limit the grasp direction to only top-down direction, we handle different objects and we incorporate a notion of reachability \cite{akinola2018workspace} to guide the grasping process. In addition, we demonstrate our method on real hardware.
Another recent work~\cite{marturi2019dynamic} looked holistically at the problem of dynamic grasping especially during handover between a human and a real robot. As the object moves, approximate inverse kinematics (IK) are computed on a database of pre-computed grasps and the quality of the IK solutions are computed and ranked. In our work, we compare the IK of filtered grasps to the current robot joint values and pick the closest grasp.


\section{Problem Definition}
The task is for a robot to pick up a moving object whose motion is not known a priori and avoid colliding with the surrounding obstacles.
We assume that the models of the objects and obstacles exist so the system can model the environment using object detection and pose estimation. The task is successful if the robot is able to pick up and lift the correct target object without knocking over the surrounding objects/obstacles. We also want target object to be picked up as fast as possible. This task imitates many warehouse conveyor belt scenarios when both the obstacle and target objects are fragile and moveable with unplanned contact.

\section{Method}

In this section we describe the various components of our system (illustrated in Figure~\ref{fig:full_pipeline}). First, we describe the visual processing unit that detects object poses. We then discuss the predictive component that estimates the future pose of the object. Next, we discuss the online grasp planning component that produces motion-conditioned reachable and stable grasps in real time. Finally, we present our arm motion generation method.

\subsection{Overview}
The overall algorithm is presented in Algorithm~\ref{alg:dynamic_grasping_alg}. Each grasp consists of a grasp pose and a pregrasp pose generated by backing off the grasp pose for distance $b$. Our pipeline takes in a known object $O$. It first retrieves a pre-computed database of grasps $G_{DB}$ for the target object; grasps in $G_{DB}$ are all in object frame. In the dynamic grasping loop, it estimates the current pose $p_c$ of the target and predicts a future pose $p_f$ with duration $t$. $t$ is defined as a step function of the euclidean distance $d$ from the arm end-effector to the planned pregrasp: $t=2\si{s}$ if $d > 0.3\si{m}$, $t=1\si{s}$ if $0.1\si{m} < d \leq 0.3\si{m}$, and $t=0\si{s}$ if $d \leq 0.1\si{m}$. We convert the grasps in $G_{DB}$ from object frame to robot frame according to predicted pose $p_f$ and then filtered the grasps using the reachability and motion-aware ranking functions described later in Section~\ref{grasp_planning} and keep the shortlisted top 10 grasps $G_F$. We pick the grasp $g_c$ from $G_F$ that is closest to the current robot configuration and move the arm if $p_c$ is reachable otherwise we continue to the next loop. We keep executing the algorithm until the condition for executing the grasp is satisfied. Define the euclidean distance between the end-effector position and the planned grasp position to be $d_p$, and the absolute quaternion distance between the end-effector orientation and planned grasp pose orientation to be $d_q$, then \textsc{CanGrasp} returns true if $d_p \leq 1.1b$ and $d_q \leq \ang{20}$, where $b$ is the back-off distance. 

After the condition is met, we get an updated estimate of the object pose, predict with horizon $t^\prime = 1\si{s}$, and convert $g_c$ using the newly predicted pose $p^\prime_f$. The arm is then moved to $g_c$. The hand is closed while moving with the target for another $t^{\prime\prime} = 0.1\si{s}$. In our pipeline, $t$, $t^\prime$, and $t^{\prime\prime}$ are configured experimentally but the optimal prediction horizon should be a function of the end-effector speed, the distance between the end-effector and the planned grasp pose, and the motion of the target. We leave this for future research. Finally, we check if the object has been lifted to determine success.

\begin{algorithm}
\footnotesize
\begin{algorithmic}[1]
\caption{Dynamic Grasping Pipeline}
\Function{DynamicGrasp}{$O$}
    \State $G_{DB}$ $\leftarrow$ \textsc{RetrieveGraspDatabase($O$)}
    \While{True}
        \State $p_c$ $\leftarrow$ \textsc{DetectPose}($O$)
        \State $p_f$ $\leftarrow$ \textsc{Predict}($p_c$, $t$)
        \State $G_W$ $\leftarrow$ \textsc{ConvertGrasps}($G_{DB}$, $p_f$)
        \State $G_F$ $\leftarrow$ \textsc{FilterGrasps}($G_{W}$, $p_f$)
        \State $g_c$ $\leftarrow$ \textsc{PickGrasp}($G_F$)
        \State Continue to next iteration if $g_c$ is not reachable
        \State Move arm to $g_c$
        \If{\textsc{CanGrasp}()}
            \State $p_c^\prime$ $\leftarrow$ \textsc{DetectPose}($O$)
            \State $p_f^\prime$ $\leftarrow$ \textsc{Predict}($p_c^\prime$, $t^\prime$)
            \State $g_c$ $\leftarrow$ \textsc{ConvertGrasps}($g_c$, $p_f^\prime$)
            \State Move arm to $g_c$ 
            \State Close hand while moving with the target for $t^{\prime\prime}$
            \State Break loop
        \EndIf
    \EndWhile
    \State \Return \textsc{CheckSuccess}()
\EndFunction
\label{alg:dynamic_grasping_alg}
\end{algorithmic}
\end{algorithm}

\subsection{Object Motion Modelling}
Picking up moving objects requires instantaneously detecting the relevant objects in the scene. We continuously track/model the motion of the object to be able to handle cases where the motion profile changes with time.

\subsubsection{Object Detection and Tracking}
In real-world experiments of this work, we use a recent learning-based method (DOPE~\cite{tremblay2018deep}) to get instantaneous poses of moving objects in the scene. DOPE trains a neural network model that takes an RGB image as input and outputs the pose of a target object relative to the camera frame. A different model is trained for each object of interest and each model can detect multiple instances of their target object. Images of the grasping scene are captured using a kinect and passed through the DOPE models to detect objects and obstacles in the scene. To achieve robustness, we use the published model~\cite{tremblay2018deep} that was trained on data collected in different lighting condition. In simulation, we directly access the pose of the objects and obstacles in the grasping scene.

\subsubsection{Recursive State Estimation\,/\,Object Pose Prediction}
Grasp\,/\,motion planning has a time cost and a computed grasp\,/\,motion plan can become obsolete very quickly as the object moves. As a result, an ability to predict the future pose of the target object can improve the overall success of a dynamic grasping system.
The motion prediction ability is needed for both planning a grasp and executing a grasp (the approach and close motion). 
While Kalman filtering (KF) \cite{kalman1960new} is a practical approach for linear motion prediction, we adopt a recurrent neural network (RNN) approach to be able to generalize to non-linear motions as well. The RNN continuously takes in a sequence of instantaneous pose measurements  $(p_{t-n}, \cdots p_{t-1}, p_{t})$ to update it's internal state which is used to predict future pose at different prediction horizon lengths ($p_{t_{f1}}, p_{t_{f2}}, \cdots, p_{t_{fm}}$).
To train the RNN \footnote{\label{rnn_footnote}RNN model: LSTM(100), 2$\times$ Dense(100), Dense(output\_shape). \\output\_shape = num\_future $\times$ dim}, we create a dataset contains planar linear, circular and sinusoidal sequence of waypoints (2000 each) randomly generated along different directions with different start points. To aid learning and generalization, each sequence data point is normalized to the start of the sequence i.e. ($(0, \cdots, p_{t-1}-p_{t-n}, p_{t}-p_{t-n}) \rightarrow (p_{t_{f1}}-p_{t-n}, p_{t_{f2}-p_{t-n}})$).
This RNN approach can also be used to model object motion during human-robot handovers.

\subsection{Grasp Planning for Moving Objects}\label{grasp_planning}
\subsubsection{Grasp Database Generation}
To generate grasps for moving objects, we pre-compute a database of grasps for all target objects while they remain static. Similar to~\cite{eppner2019billion}, this database was collected and evaluated purely in simulation with dynamics turned on.
First, we densely generate 5000 stable grasps for each object using a simulated annealling approach~\cite{ciocarlie2009hand}, and we then evaluate all the grasps in simulation; each grasp is executed to lift the object 50 times and each time we add random noise to the object pose~\cite{weisz2012pose}. The success rate gives a measure of the robustness of the grasp and we choose the top 100 robust grasps.

\subsubsection{Reachability-Aware Grasping}
In the dynamic grasping setting, it is important to have a fast way to choose a feasible grasp out of the list of stable and robust grasps. Generating collision-free IK for all the grasps can be time-consuming; instead, we use our pre-computed reachability space to quickly rank the grasps for the given object pose estimate (See \cite{akinola2018workspace} for more details). The larger reachability value the grasp has, with higher probability a valid IK can be found for that grasp. Reachability can also be an index of manipulability and it follows the intuition that the most reachable grasp has higher probability to continue being reachable in the future, reducing the number of grasp switches while the target moves around. The interpolation and indexing of a pre-computed 6D space gives a fast way to reduce the grasp database to a few more reachable grasps whose IK can then be found. This approach is much faster that computing IK for the entire database and is important in dynamic settings. 
We can use this reachability computation as a rank function for \textsc{FilterGrasp} in Algorithm~\ref{alg:dynamic_grasping_alg}. An example using reachability is shown in Figure~\ref{fig:full_pipeline}. 

\subsubsection{Motion-Aware Grasping}
We observe that the success rate of a stable grasp varies depending on the motion of the object. For example, picking up an object from behind as it moves away can result in different success rate statistics compared to approaching in the direction opposite it's motion.
To address this, we learn a neural network model $\mathcal{M}(g, pg, v, \theta)$ that predicts the success probability of a grasp $g$ given the motion profile of the object (speed $v$ and motion direction $\theta$). The input into the model includes:
\begin{itemize}
    \item The 6D grasp pose ($g \in  \mathbb{R}^6$) i.e. the $\{x, y, z\}$ position and $\{roll, pitch, yaw\}$ orientation  in the object's frame of reference.
    \item The 6D pre-grasp pose ($pg \in  \mathbb{R}^6$) which is the grasp pose backed off (5 cm for Mico hand and 7.5cm for robotiq hand) along the approach direction (i.e. a vector pointing from the end-effector towards the object).
    \item The speed $v \in \mathbb{R}$ of the object.
    \item The motion direction $\theta \in [0, 2\pi]$. We assume a 2D planar motion parameterized by a polar angle direction around the z-axis of the object frame.
\end{itemize}
The model has two hidden layers (512 each) and an output predicting the success probability. We generated a dataset of 10000 grasp attempts each on 7 different objects in simulation using the robot's end-effector only. For each grasp attempt, the end-effector starts at the pregrasp pose and moves towards the object while the object moves in a randomly sampled planar direction, at a speed sampled uniformly between 0.5\si{cm/s} and 5\si{cm/s}. We record the result of the grasp attempts and use this as supervision to train the models (one for each object). We train 100 epochs for each object. The average training time is $\sim5$\si{min}s and the average validation accuracy is $0.963$ with False Positive Rate (FPR) 0.017 and False Negative Rate (FNR) 0.117. Ultimately, the probability of success output by the network can be used as a motion-aware quality conditioned on the object motion. We can use this network to quickly filter grasps that has the highest motion-aware quality for \textsc{FilterGrasp} in Algorithm~\ref{alg:dynamic_grasping_alg}. 
In general, the motion-aware model prefers grasps facing the moving direction of the target.

\subsubsection{Combining Reachability and Motion-aware}\label{sec:combine}
We want to include grasps in the shortlisted pool $G_F$ that are both reachable and are stable conditioned on the object motion. There are many different ways to combine the reachability and motion-aware quality for each grasp in the database. We empirically find that simply including the top 5 grasps with highest reachability and the top 5 grasps with highest motion-aware quality outperforms other ways of combination, including the weighted sum of two values or filtering by reachability and then motion-aware quality, etc.

\begin{figure*}
\vspace{2mm}
\centering
 \begin{subfigure}{0.195\linewidth}
    \centering
    \includegraphics[width=0.975\textwidth]{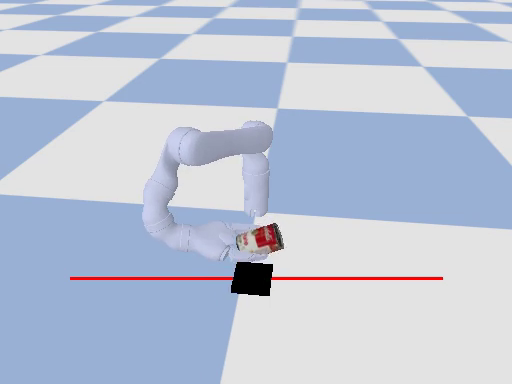}
	\caption{\small Linear}
	\label{fig:linear_no_obstacles}
\end{subfigure}
 \begin{subfigure}[h]{0.195\linewidth}
    \centering
    \includegraphics[width=0.975\textwidth]{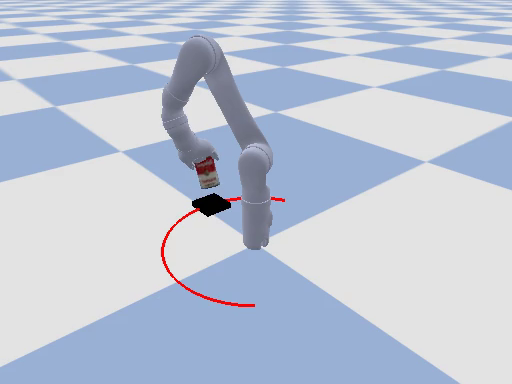}
	\caption{\small Circular}
	\label{fig:circular}
\end{subfigure}
 \begin{subfigure}[h]{0.195\linewidth}
    \centering
    \includegraphics[width=0.975\textwidth]{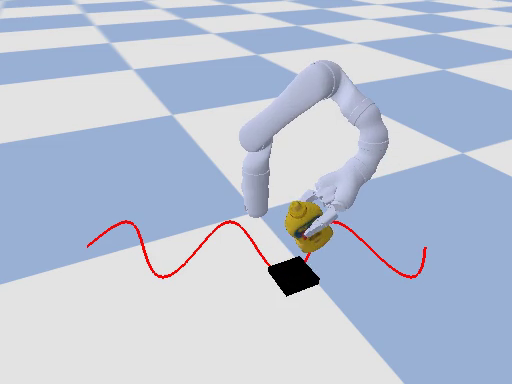}
	\caption{\small Sinusoidal}
	\label{fig:sinusoidal}
\end{subfigure}
 \begin{subfigure}{0.195\linewidth}
    \centering
    \includegraphics[width=0.975\textwidth]{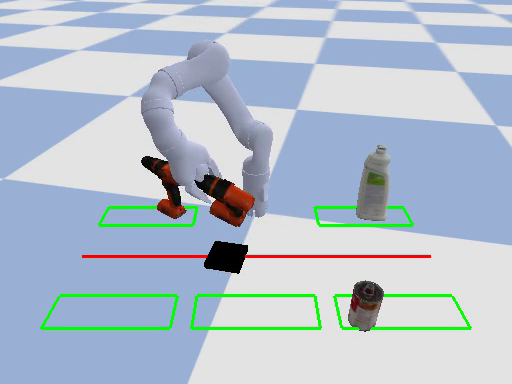}
	\caption{\small Static  Obstacles}
	\label{fig:linear_static_obstacles}
\end{subfigure}
 \begin{subfigure}{0.195\linewidth}
    \centering
    \includegraphics[width=0.975\textwidth]{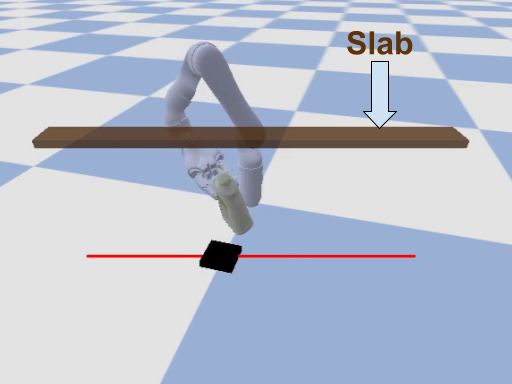}
	\caption{\small Slab Fixture}
	\label{fig:linear_slab_fixture}
\end{subfigure}
 \begin{subfigure}{\linewidth}
    \centering
    \includegraphics[width=0.19\textwidth]{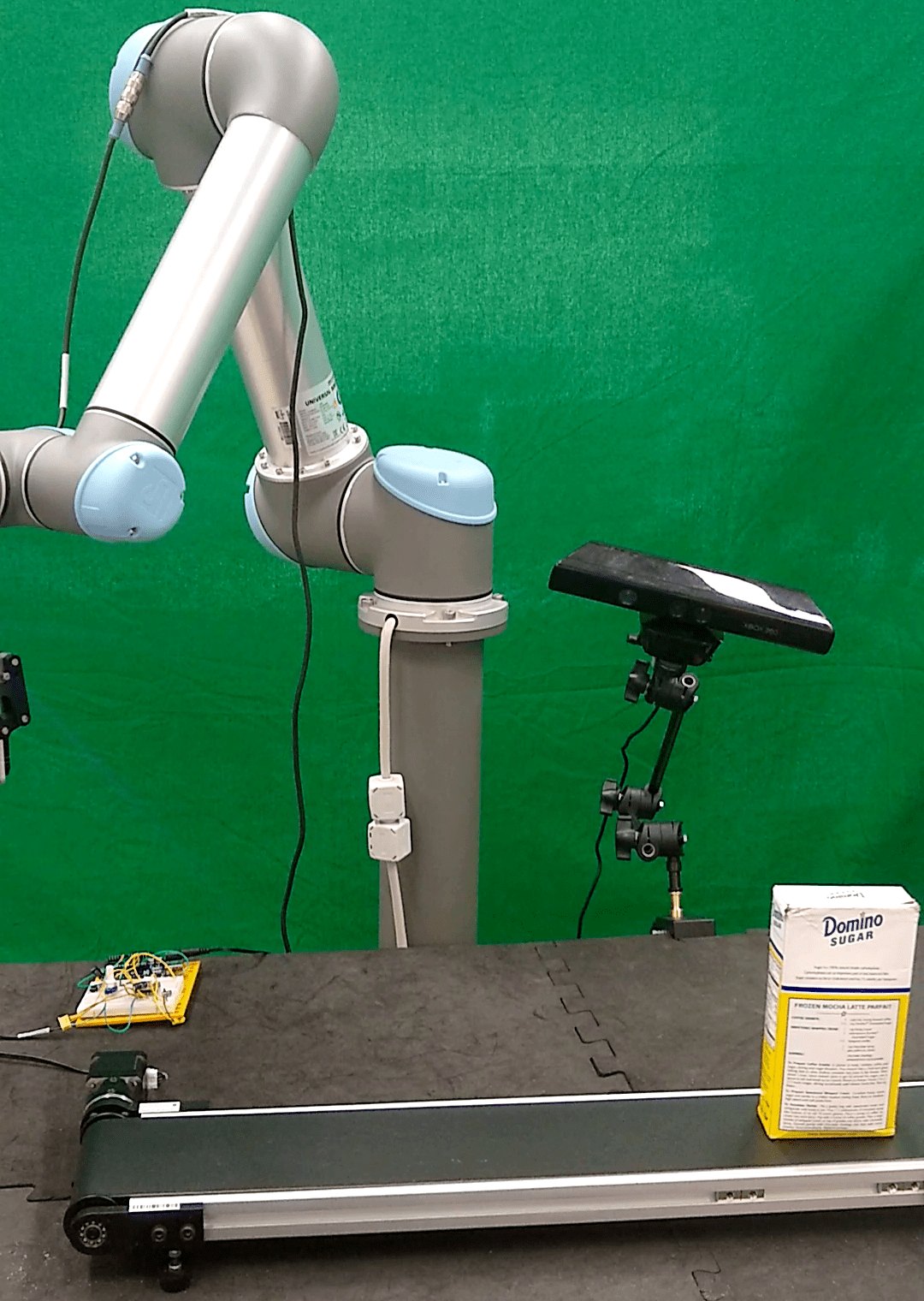}
    \hfill
    \includegraphics[width=0.19\textwidth]{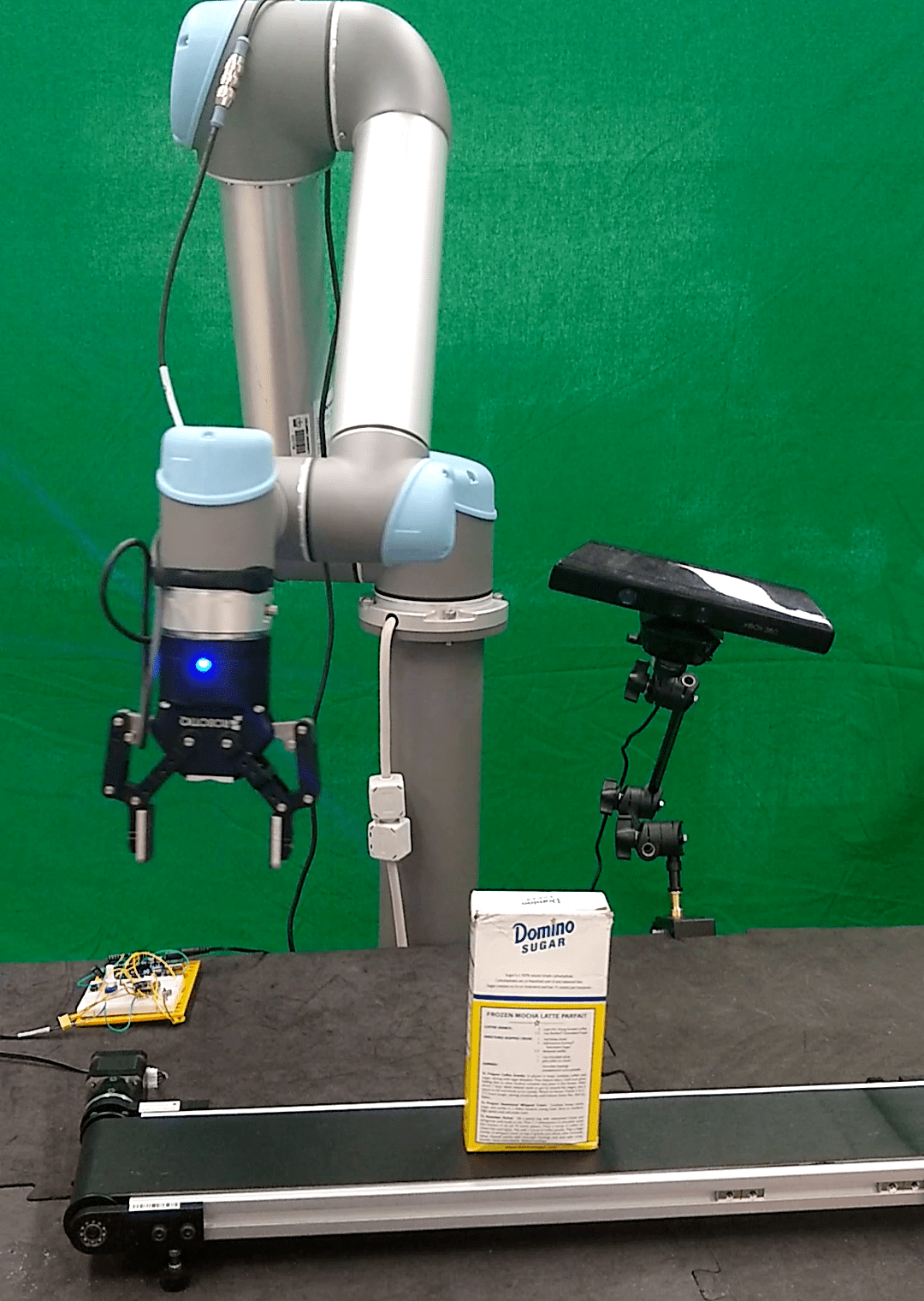}
    \hfill
    \includegraphics[width=0.19\textwidth]{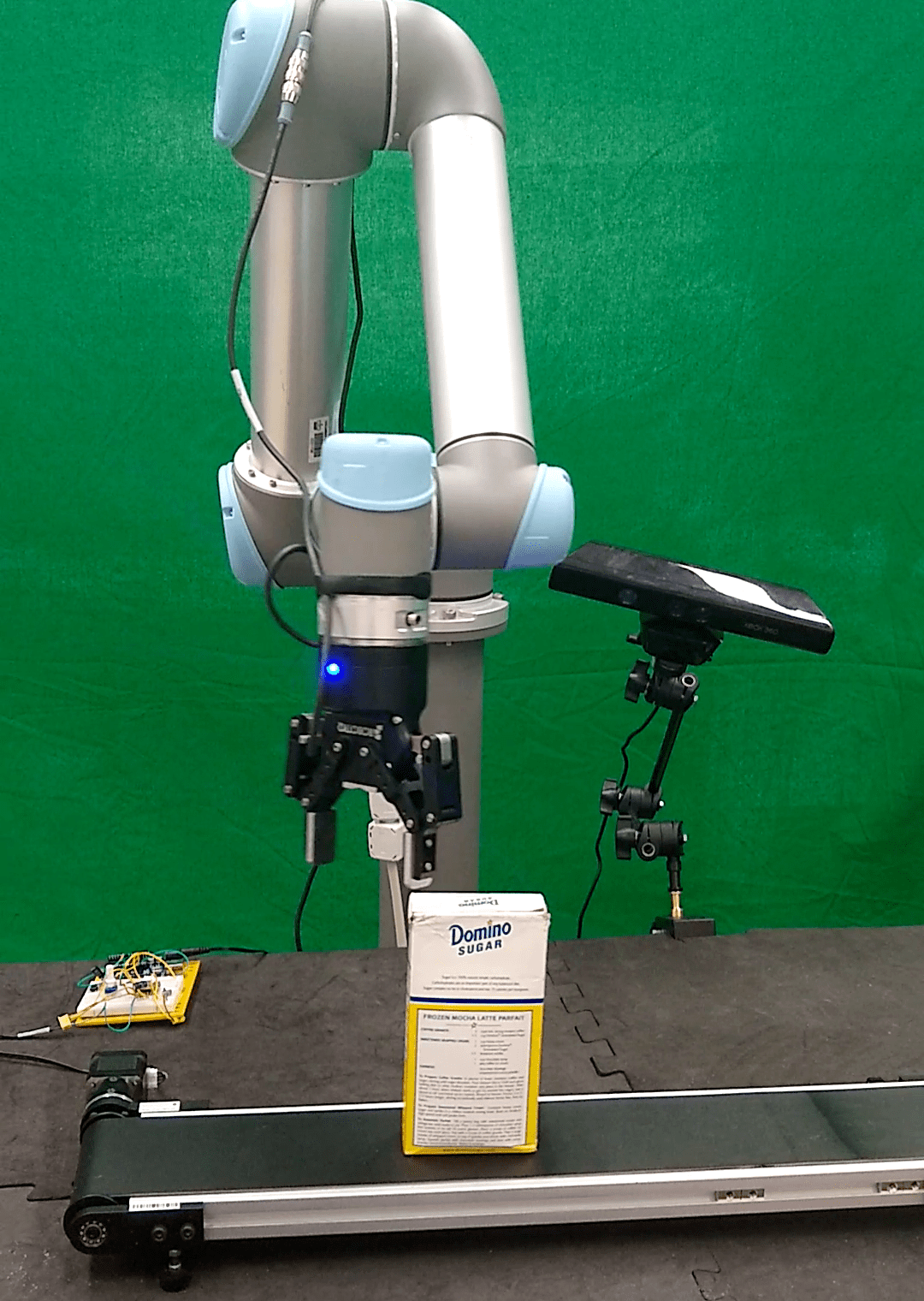}
    \hfill
    \includegraphics[width=0.19\textwidth]{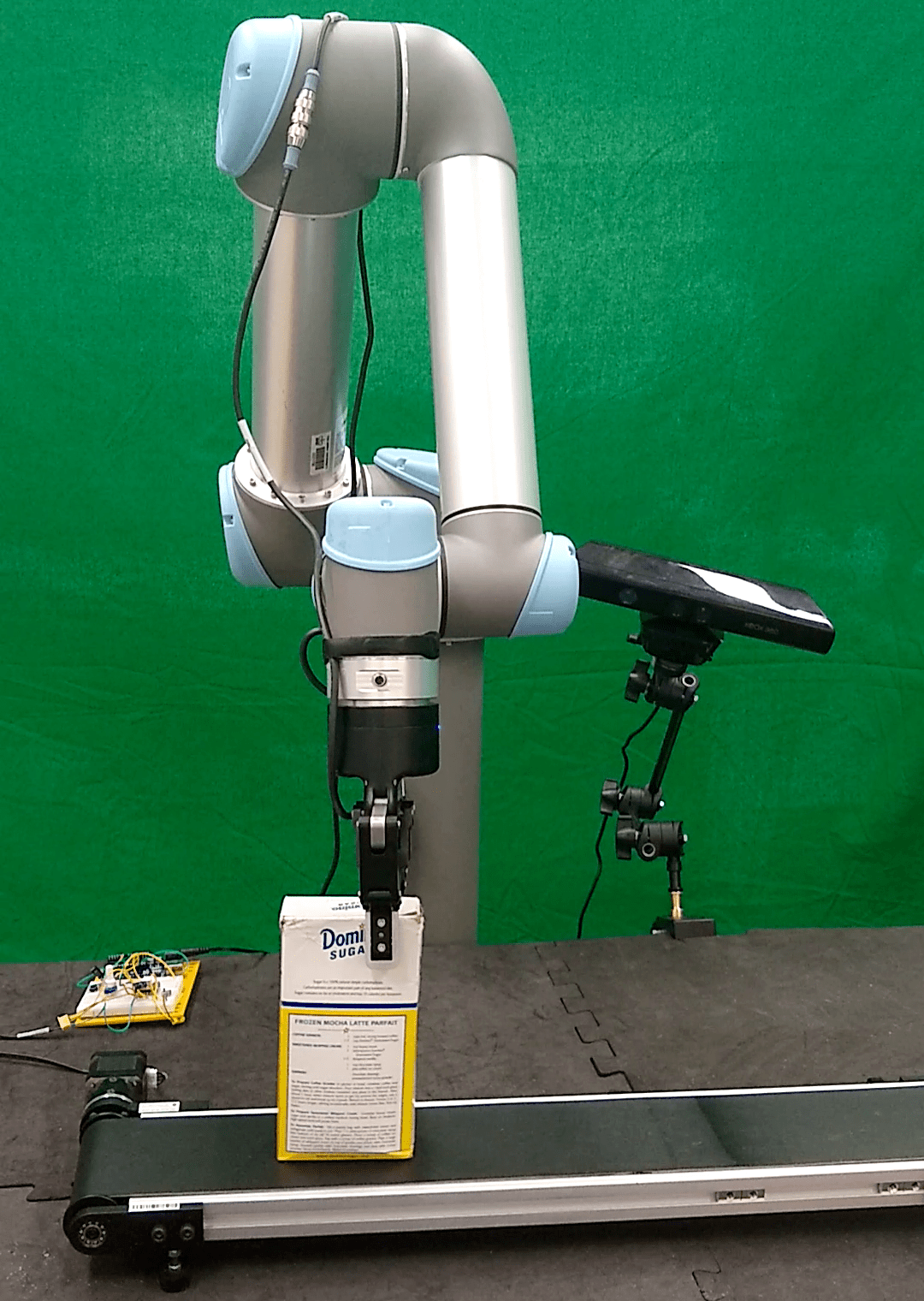}
    \hfill
    \includegraphics[width=0.19\textwidth]{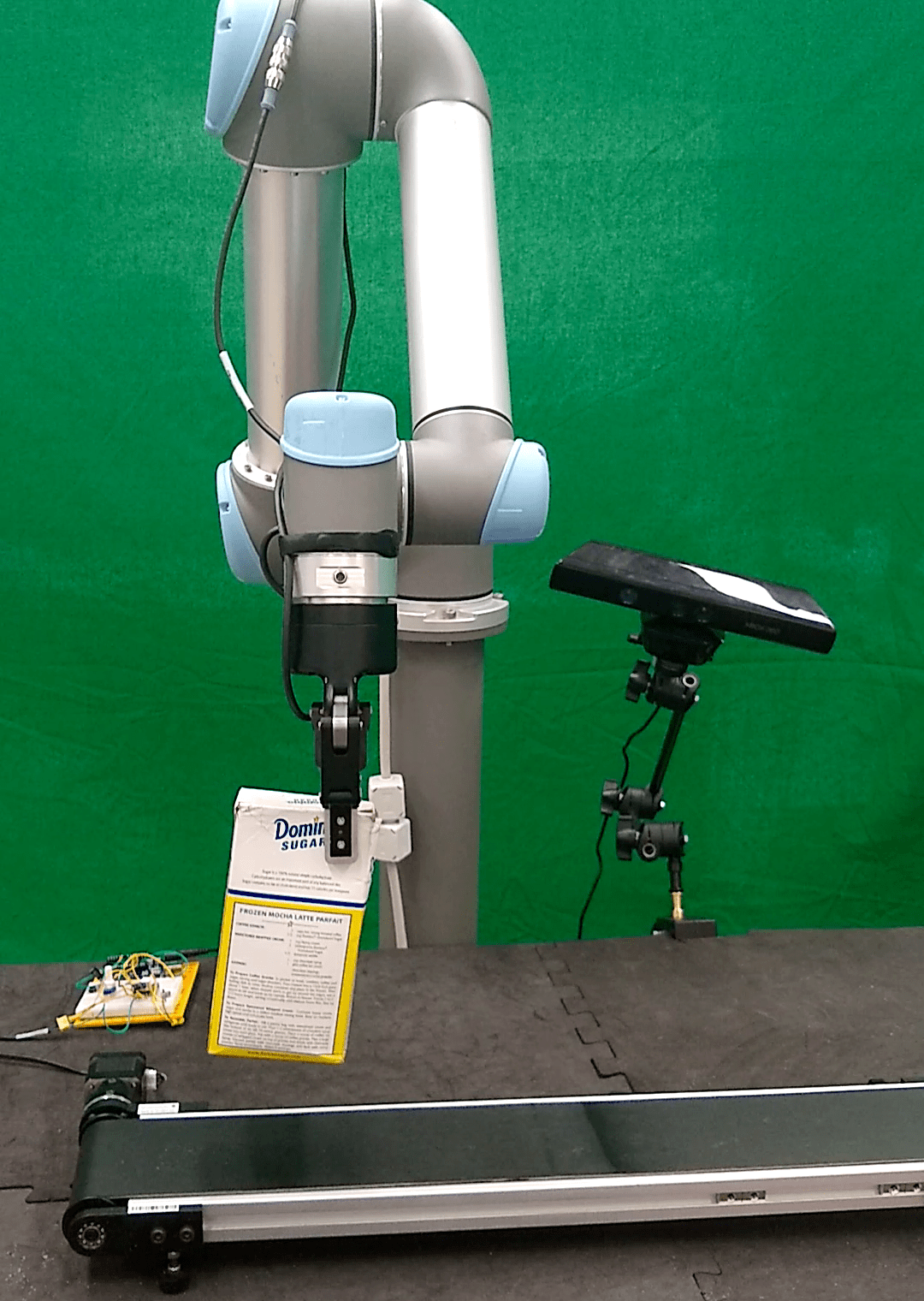}
	\caption{\small Real Robot }
	\vspace{-2mm}
    \label{fig:linear_real_robot}
\end{subfigure}

\caption{\small Dynamic Grasping Tasks. Experimental scenarios for picking up objects on a conveyor belt. The red line shows the conveyor belt trajectory.
\textbf{(a), (b), (c)} Linear, circular and sinusoidal motion of target object with no surrounding obstacles.
\textbf{(d)} Linear motion with surrounding static obstacles. Green rectangles are the sub-regions where we sample obstacle locations.
\textbf{(e)} Linear motion with slab fixture that limits feasible grasping directions.
\textbf{(f)} Real Robot Demo: Linear motion of target object moving at 4.46 cm/s.}
\label{fig:dynamic_grasping_task_execution}
\vspace{-5mm}
\end{figure*}

\subsection{Motion Generation and Grasp Execution}


There are three stages of motion generation when picking up an object: reaching, grasping and lifting~\cite{menon2014motion}. In our implementation, we transform the planned grasp to match the predicted future pose of the object and generate arm motion for all three phases.

To be able to generate and update the reaching trajectories as the object moves, we introduce the idea of trajectory seeding that uses the trajectory solution of a previous time step as an initialization for finding a new trajectory at the current time-step.
For sampling-based methods like RRT or PRM, this entails initializing the sampling tree or roadmap with the waypoints found from the previous time step. This seeds the search to be quite close to the previous path and empirically helps find new solutions that are not drastically different from the previous solution. An unconstrained sampling based approach can return drastically different trajectories in subsequent trajectories which can be disadvantageous in dynamic settings. Another benefit of our seeding approach is that a good initialization from seeding can reduce the time used to find a valid solution.
We implement our seeding approach on CHOMP, RRT and PRM and find that PRM works best for our experimental tasks.

Note that the motion plan is executed once generated interrupting the previous trajectory that was being executed.
To ensure that the arm does not slow down as new trajectories are computed and updated, we retime the trajectory solution from the solver so that it blends with the current arm velocities and it moves at fast as possible while also respecting joint limits~\cite{bobrow1985time}.
We use cartesian control to move the arm during the grasping and lifting stages.

\section{Experiments}
We extensively evaluate the performance of our algorithm picking different target objects in randomized linear\,/\,nonlinear motion with\,/\,without static obstacles in simulation. We then demonstrate that our method works reliably on a real robot. Videos showing some of the experiments can be found at our project website \url{http://crlab.cs.columbia.edu/dynamic_grasping}.

\subsection{Experimental Setup}
We create different scenarios illustrated below using the Bullet simulator~\cite{coumans2016pybullet} to evaluate the performance of our methods on two robot arms with parallel jaw grippers: the Kinova Mico and the UR5-Robotiq robots. These two robots have different workspace dimensions, manifolds, joint limits as well as different gripper spans\,/\,width. UR5 arm has a wider span and moves quicker than Mico arm, so we intentionally make the tasks for UR5-Robotiq harder. For each robot, we simulate the task of linear and non-linear conveyor belt pickup, which plays a significant role in warehouse packaging and assembly lines. In these scenarios, there is a target object moving on a belt, possibly among surrounding static obstacles. Both the target and the obstacle objects can be fragile and and we cannot knock them over.

\begin{figure}[!h]
\begin{center}
    \includegraphics[width=0.3\linewidth]{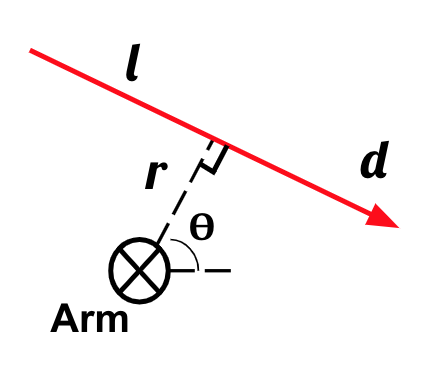}
    \includegraphics[width=0.3\linewidth]{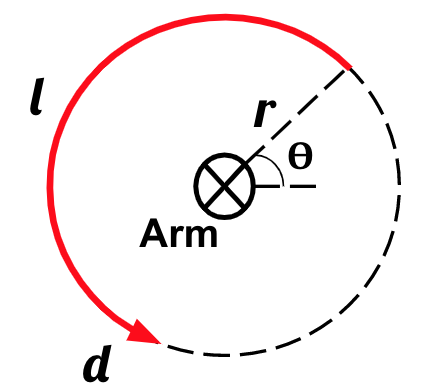}
    \includegraphics[width=0.3\linewidth]{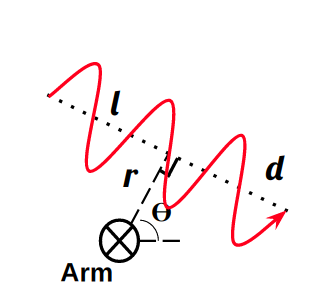}
\end{center}
\caption{ \small A bird's-eye view of randomized linear, circular and sinusoidal conveyor belt motion generation process. A random experiment motion is parameterized by angle $\theta$, distance $r$, direction $d$, and length $l$. The cross indicates the position of the robot base. The red line shows the motion of the conveyor belt, with an arrow indicating the direction. The horizontal dashed line indicates the $x$-axis of the world frame. Left: linear motion. Middle: circular motion. Right: sinusoidal motion.}
\label{fig:random_motion}
\end{figure}

\paragraph{Linear Motion} The target object moves linearly at a constant speed (3\si{cm/s} for Mico and 5\si{cm/s} for UR5-Robotiq) as shown in Figure~\ref{fig:linear_no_obstacles}. The conveyor trajectories are randomized using 4 parameters as shown in Figure \ref{fig:random_motion}. $\theta$ specifies the counter clockwise angle of the line perpendicular to the linear motion, connecting the middle point of the motion and the base of the arm. $r$ is the distance between the linear trajectory and the arm base. $l$ is the length of the linear trajectory. $d \in \{+1, -1\}$ indicates the direction of the motion, where $+1$ means moving counter clockwise and $-1$ means moving clockwise. We set $0 \leq \theta < 2\pi$ (radians), $0.15m \leq r \leq 0.4m$ for Mico and $0.3m \leq r \leq 0.7m$ for UR5-Robotiq, and $l = 1m$.
\paragraph{Linear with Obstacles} We add 3 static obstacles to the linear motion in the grasping scene. Specifically, we divide the near region (distance between 0.15\si{m} and 0.25\si{m}) surrounding the linear motion into 5 sub-regions, as shown in Figure~\ref{fig:linear_static_obstacles}. For each obstacle, we randomly pick a sub-region and uniformly sample a location in the sub-region. We make sure there is no collision between obstacle and robot or between two obstacles. The arm has to avoid hitting both obstacles and the target. 
\paragraph{Linear with Top Slab} A 2\si{cm}-thick slab of width 10\si{cm} is placed 40\si{cm} directly on top of the conveyor belt. This limits the grasping directions (e.g. top-down grasps) and makes motion planning/grasping more challenging.
\paragraph{Linear with Z Motion} We relieve the constraint of the linear motion so that the object can also move in the Z axis. The starting height and the end height is randomly sampled between 0.01\si{m} and 0.4\si{m}.
\paragraph{Linear with Varying Speed} The target is accelerated from 1\si{cm/s} to 3\si{cm/s} and from 3\si{cm/s} to 5\si{cm/s} for Mico and UR5-Robotiq respectively.
\paragraph{Circular Motion} A smooth non-linear circular motion as shown in Figure~\ref{fig:circular}. The speed of the conveyor belt is constant (2\si{cm/s} for Mico and 3\si{cm/s} for UR5-Robotiq). The circular motion trajectory is also randomized by 4 parameters as shown in Figure~\ref{fig:random_motion}. $\theta$ controls the angle of the starting position on the circle. $r$ is the radius of the circle. $l$ specifies the length of the motion. $d \in \{+1, -1\}$ indicates the direction of the motion, where $+1$ means moving counter clockwise and $-1$ means moving clockwise. We set $0 \leq \theta < 2\pi$ (radians), $0.15\si{m} \leq r \leq 0.4\si{m}$ for Mico and $0.3\si{m} \leq r \leq 0.7\si{m}$ for UR5-Robotiq, and $l = 1\si{m}$.
\paragraph{Sinusoidal Motion} This is a more challenging non-linear motion where the object moves along a sinusoidal path as shown in Figure~\ref{fig:sinusoidal}. To do this, a sinusoid is super-imposed on the randomly generated linear motion as shown in Figure~\ref{fig:random_motion}. In addition to the parameters of the linear motion ($\theta, r, l, d$), we specify the amplitude $A$ and frequency $f$ of the sinusoid. We set $A = l/8\si{m}$  and $f = 2 \pi / (l/ 3)\si{Hz}$.

\begin{figure}
\vspace{2mm}
\begin{center}
    \includegraphics[width=0.995\linewidth]{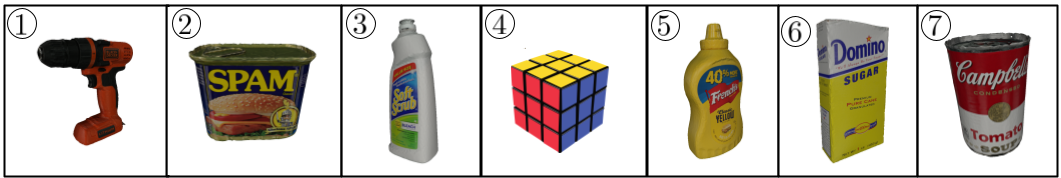}
\end{center}
\caption{\small Seven objects from  the YCB Object Database selected as the graspable objects in our experiments. All seven are used for simulation experiments while the last three are used for the real robot experimentation.}
\label{fig:ycb_objects}
\vspace{-5mm}
\end{figure}

Each experiment in simulation is run on 7 different target objects shown in Figure~\ref{fig:ycb_objects} whose sizes can physically fit in the robots' hands. The randomized process for generating conveyor belt motion ensures that the results are not biased to a specific robot configuration. For example, a particular starting pose might be close to the robot arm end-effector and will have a higher success rate. For each object, we run each 100 times and report the average success rate and grasping time across 700 trials. In each setting, we compare the performance of the below methods.
\begin{itemize}
    \item \textbf{Ours (R+M).} This is our proposed method that uses all the discussed modules and filters grasp in the grasp database combining both reachability and motion-aware quality as discussed in Section~\ref{sec:combine}.
    \item \textbf{Ours (Reachability).} Same as \textit{Ours (R+M)} except the grasps are filtered with only reachability.
    \item \textbf{Ours (Motion-aware).} Same as \textit{Ours (R+M)} except the grasps are filtered with only motion-aware quality.
    \item \textbf{Baseline.} Using randomly sampled 10 grasps from the grasp database as the filterd grasps. We need to use a subset because checking IK for all grasps from the database during dynamic grasping is unfeasible with very low success rate that is not worth comparing.  
    \item \textbf{No Traj. Seeding.} This ablation study picks the model from above with best performance and removes the trajectory seeding module to study its importance.
    \item \textbf{No Prediction.} Similar to \textit{No Traj. Seeding}, this removes the prediction module to study its importance.
\end{itemize}
We evaluate a subset of the experiments on a real robot hardware to validate our approach. For this, we use the UR5 robot arm fitted with a Robotiq parallel jaw gripper to pick up an object moving on a conveyor belt.

\begin{table*}[t]
  \small
  \centering
  \vspace{0.06in}
  \caption{
  \small Simulation Experiments for the Kinova Mico (Top) and UR5 (Bottom) robot arms. For each entry, run on 7 objects and 100 trials each. We report success rate, dynamic grasping time (\si{s}) averaged over 700 trials. 
}
  \resizebox{\textwidth}{!}{
  \begin{tabular}{c|ccccc|c|c}
    \toprule
    Methods & \begin{tabular}[c]{@{}c@{}}Linear~(3cm/s)\end{tabular} & \begin{tabular}[c]{@{}c@{}}Linear~(3cm/s)~\\with~Obstacles\end{tabular} & \begin{tabular}[c]{@{}c@{}}Linear~(2cm/s)~\\with~Top~Slab\end{tabular} & \begin{tabular}[c]{@{}c@{}}Linear~(3cm/s)~\\with~Z~Motion\end{tabular} &
    \begin{tabular}[c]{@{}c@{}}Linear~(1-3cm/s)~\\Varying Speed\end{tabular} & \begin{tabular}[c]{@{}c@{}}Circular\\(2cm/s)\end{tabular} & \begin{tabular}[c]{@{}c@{}}Sinusoidal\\(1cm/s)\end{tabular}  \\
    \midrule
    Ours (R + M)   & 0.799, 10.23s & \textbf{0.796, 11.43s} & \textbf{0.781,  21.92s} & \textbf{0.814, 10.39s} & \textbf{0.869, 9.988s} & \textbf{0.875, 10.38s} & \textbf{0.895, 8.661s} \\
    Ours (Reachability)   & \textbf{0.802, 10.21s} & 0.795, 10.13s & 0.759, 19.91s & 0.806, 9.836s & 0.836, 9.609s & 0.861, 9.421s & 0.867, 7.857s \\
    Ours (Motion-aware)   & 0.722, 15.30s & 0.780, 14.37s & 0.697, 25.07s & 0.710, 14.61s & 0.819, 14.45s & 0.857, 15.43s & 0.827, 13.19s \\   
    Baseline      & 0.439, 23.85s & 0.419, 24.15s & 0.431, 38.12s & 0.430, 23.19s & 0.610, 25.72s & 0.716, 24.89s & 0.643, 20.22s \\
    \midrule
    \midrule
    No Traj. seeding    & 0.769, 10.61s & 0.777, 10.83s & 0.706, 22.29s & 0.800, 10.41s & 0.827, 10.23s & 0.857, 10.01s & 0.827, 8.588s \\
    No Prediction      & 0.610, 12.91s & 0.614, 13.30s & 0.609, 25.84s & 0.737, 12.60s  & 0.761, 11.96s & 0.767, 13.11s & 0.807, 9.045s\\
    \bottomrule
  \end{tabular}
  \label{tab:main}
  }
\end{table*}

\begin{table*}[t]
  \small
  \centering
  \resizebox{\textwidth}{!}{
  \begin{tabular}{c|ccccc|c|c}
    \toprule
    Methods & \begin{tabular}[c]{@{}c@{}}Linear~(5cm/s)\end{tabular} & \begin{tabular}[c]{@{}c@{}}Linear~(5cm/s)~\\with~Obstacles\end{tabular} & \begin{tabular}[c]{@{}c@{}}Linear~(3cm/s)~\\with~Top~Slab\end{tabular} & \begin{tabular}[c]{@{}c@{}}Linear~(5cm/s)~\\with~Z~Motion\end{tabular} &
    \begin{tabular}[c]{@{}c@{}}Linear~(3-5cm/s)~\\Varying Speed\end{tabular} & \begin{tabular}[c]{@{}c@{}}Circular\\(3cm/s)\end{tabular} & \begin{tabular}[c]{@{}c@{}}Sinusoidal\\(1cm/s)\end{tabular}  \\
    \midrule
    Ours (R + M)   & \textbf{0.874, 8.134s} &  \textbf{0.854, 9.104s} & \textbf{0.748, 19.86s} & 0.858, 8.166s  & \textbf{0.917, 7.895s} & \textbf{0.909, 8.311s} & \textbf{0.946, 9.243s} \\
    Ours (Reachability)   & 0.857, 8.730s & 0.841, 9.646s & 0.652, 18.52s & \textbf{0.872, 8.864s} & 0.907, 8.748s & 0.890, 9.512s & 0.925, 8.608s \\
    Ours (Motion-aware)   & 0.744, 9.976s & 0.752, 9.896s & 0.675, 19.86s & 0.717, 10.47s & 0.854, 8.935s & 0.840, 10.68s & 0.930, 9.645s \\   
    Baseline   & 0.676, 12.64s & 0.576, 13.63s & 0.606, 22.89s & 0.659, 12.65s & 0.788, 12.84s & 0.810, 14.09s & 0.716, 17.19s \\
    \midrule
    \midrule
    No Traj. seeding   & 0.849, 9.107s & 0.810, 10.22s & 0.631, 20.01s & 0.836, 9.047s & 0.906, 8.859s & 0.899, 9.610s &  0.904, 11.17s  \\
    No Prediction   & 0.284, 10.31s & 0.269, 11.41s & 0.457, 20.04s & 0.261, 10.89s & 0.344, 10.44s & 0.594, 9.891s & 0.310, 11.96s \\
    \bottomrule
  \end{tabular}
  }\label{tab:ur5_results}
\end{table*}

\subsection{Experimental Results and Discussion}
\label{sec:conveyor_belt_experimental_results}
Shown in Table~\ref{tab:main}, our proposed methods with reachability and motion awarenesses or a combination of both outperform the baseline in all cases. The ablation studies demonstrate the importance of the trajectory seeding and motion prediction components.
\subsubsection{Effect of Grasp Planning}
The results show that reachability and motion awarenesses help extenssively in dynamic grasping tasks. This is because our methods leverage these two ranking functions to quickly filter grasps that are likely to be reachable or stable conditioned on the current motion of the object. They help to focus on those grasps which are near-optimal given the object pose and motion and reduce unnecessary IK calls. Without reachability or motion awareness, even though all grasps in the grasp database are stable in static cases, the selected 10 grasps might not be as optimal given the current location where the object has moved or the current motion the object is following. 

\begin{figure}
\begin{center}
    \includegraphics[width=0.99\linewidth]{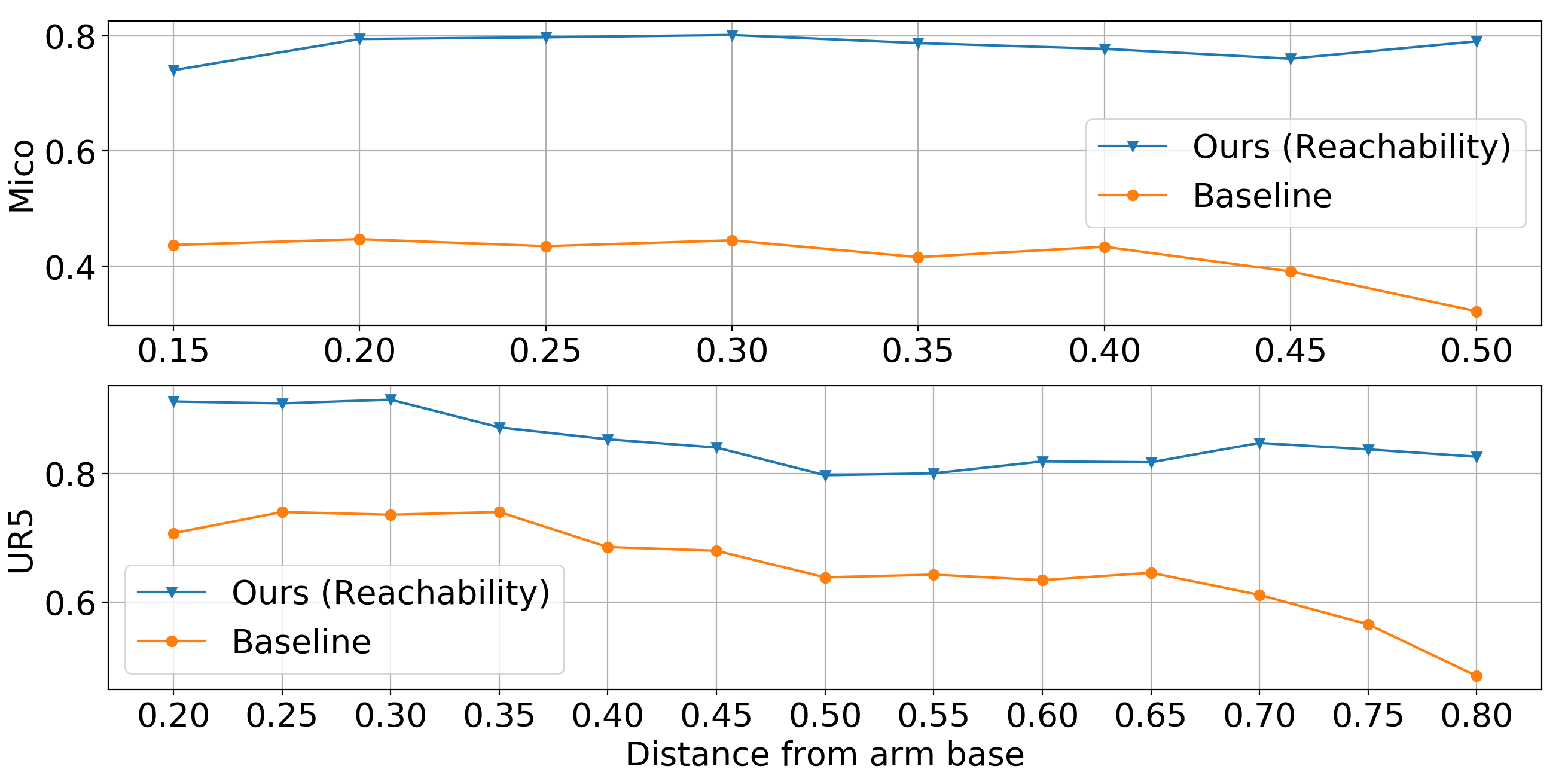}
\end{center}
\caption{\small Success rate vs.\ distance. Improvement from reachability awareness becomes more significant when the moving object is extremely close to or far from the robot.}
\label{fig:reachability_vs_distance}
\vspace{-5mm}
\end{figure}

We also notice that \textit{Ours (Reachability)} almost always performs better than \textit{Ours (Motion-aware)}. This shows that in dynamic grasping setting, reachability awareness can be a slightly more important factor than motion-aware. Though being stable for the motion of the object, the selected grasps by motion-aware might still be unreachable and waste some IK computing time. We further investigate the relationship between reachabiity performance and the distance of the conveyor motion to the robot base, using linear motion while varying $r$, as demonstrated in Figure~\ref{fig:reachability_vs_distance}. We find that reachability being especially beneficial in difficult-to-reach near and far portions of the workspace.

\textit{Ours (R + M)} outperforms \textit{Ours (Reachability)} in all cases except linear motion for Mico and linear with Z motion for UR5 arm. Combining reachbility and motion awareness include grasps that are both reachable and robust for the current motion. This combines the advantages of both methods and provides a pool of wider variety for \textsc{PickGrasp} to choose from. For example, even though a grasp might not be the most reachable, but it can also be included in the filtered grasps because of high motion-aware quality. For the two cases where \textit{Ours (Reachability)} outperforms \textit{Ours (R+M)}, we believe it is because the most motion-aware grasps happen to have very low reachability but are closer to the current robot configuration. They are picked but cannot remain reachable and result in unnecessary grasp switches.

\subsubsection{Effect of Seeding in Arm Trajectory Generation}
We observe that the value of seeding during trajectory generation becomes significant for tasks when slab fixture is above the conveyor belt limiting the range of motion of the arm (column 3 of Table~\ref{tab:main}, Mico and UR5 show a performance drop of 7.5\% and 11.7\% respectively), and when the motion is sinusoial and hard to model (column 7 of Table~\ref{tab:main}, Mico and UR5 show a performance drop of 6.8\% and 4.2\% respectively). Without seeding, computing arm trajectory from scratch at each time step is computationally expensive. Besides, seeding makes the new trajectory solution similar to the previous one. Qualitatively, we noticed that seeding makes the arm motion less wavy given that the arm trajectories generated in subsequent time steps is seeded to be similar to the immediate previous one.

\subsubsection{Effect of Object Motion Prediction} 
Our results also show that object motion prediction is an important component for dynamic grasping and we see the biggest drop in performance without motion prediction. This is expected as we can image without prediction in dynamic grasping, even with perfect grasp planning and optimal motion generation, the gripper will never catch the target because of the delay from computation. Its importance becomes more pronounced as the object's motion is hard to predict (non-linear\,/\,varying speed) and also when the gripper width is smaller. We observe that there is a bigger drop in performance for the UR5-Robotiq robot, compared to the Mico robot. Qualitatively, we observe that a lot of the failure cases occur during the approach-and-grasp phase where the robot finger narrowly knocks off the object. The wider span of the Mico gripper enables it to be more robust in this sense.

\subsubsection{Real Robot Demonstration}
We demonstrate our algorithm on the real robot by picking up objects 5, 6, 7 shown in Figure~\ref{fig:ycb_objects} as each object moves on a conveyor belt with no surrounding obstacles. We repeat this experiment 5 times and the success rates for objects  5, 6, 7 are $4/5$, $5/5$ and $3/5$ respectively. Even though the object is moving relatively fast (4.46 cm/s), our method is able to pick the objects 5 and 6 reliably well. The robot is able to align its gripper along the narrow axis of the objects and pick them up while moving. The failure cases for object 7 is because the radius of the tomato can is slightly smaller than the gripper span with a tight margin for error in the approach and grasp stage.

%

\section{Conclusion}
This work presents a novel pipeline for dynamic grasping moving objects with reachability and motion awareness. We demonstrate its ability with a RNN motion predictor and adapive motion planning with seeding. We show in experiments with various settings that these elements are important to improve the performance.
This work is a model-based visual-pose feedback system. A future work will be the image-based analogue where arm-hand trajectory commands are directly generated based on image\,/\,depth image features using learning-based techniques.










\end{document}